%% file: main.tex
\documentclass[12pt]{iopart}
\usepackage[ruled,vlined]{algorithm2e}

\usepackage{iopams}  
\usepackage{xcolor}
\usepackage{graphicx}
\usepackage{subcaption}
\usepackage{bigdelim, array}
\usepackage{multicol}
\usepackage{multirow}
\usepackage[toc,page]{appendix}

\begin{document}

\title[]{Geometric GNNs for Charged Particle Tracking at GlueX}

\author{Ahmed Hossam Mohammed$^{*, 1}$, Kishansingh Rajput$^{1, 2}$, Simon Taylor$^1$, Denis Furletov$^3$, Sergey Furletov$^1$, Malachi Schram$^{1,4}$}
\address{$^1$ Thomas Jefferson National Accelerator Facility, Newport News, VA 23606, USA}
\address{$^2$ Department of Computer Science, University of Houston, Houston, TX 77204, USA}
\address{$^3$ Department of Physics, William \& Mary, Williamsburg, VA 23185, USA}
\address{$^4$ Computer Science Department, Old Dominion University, Norfolk, VA 23529, USA}
\ead{$^*$ ahmedm@jlab.org}
\vspace{10pt}
\begin{indented}
\item[]May 2025
\end{indented}

\begin{abstract}
Nuclear physics experiments are aimed at uncovering the fundamental building blocks of matter. 
The experiments involve high-energy collisions that produce complex events with many particle trajectories. 
Tracking charged particles resulting from collisions in the presence of a strong magnetic field is critical to enable the reconstruction of particle trajectories and precise determination of interactions. 
It is traditionally achieved through combinatorial approaches that scale worse than linearly as the number of hits grows. 
Since particle hit data naturally form a 3-dimensional point cloud and can be structured as graphs, Graph Neural Networks (GNNs) emerge as an intuitive and effective choice for this task. 
In this study, we evaluate the GNN model for track finding on the data from the GlueX experiment at Jefferson Lab. 
We use simulation data to train the model and test on both simulation and real GlueX measurements. 
We demonstrate that GNN-based track finding outperforms the currently used traditional method at GlueX in terms of segment-based efficiency at a fixed purity while providing faster inferences. 
We show that the GNN model can achieve significant speedup by processing multiple events in batches, which exploits the parallel computation capability of Graphical Processing Units (GPUs). 
Finally, we compare the GNN implementation on GPU and FPGA and describe the trade-off.
\end{abstract}

\input{Introduction}
\input{Previous-Work}
\input{Background}
\input{Pre-Processing}
\input{Modeling}
\input{Results}
\input{Conclusion}
\input{Acknowledgements}

\section*{References}
\bibliographystyle{iopart-num}
\bibliography{references}

\newpage
\input{Appendix}
\end{document}

%% file: Introduction.tex
\section{Introduction}
\label{sec:Introduction}

% Descripton of particle propagation in detectors
Nuclear Physics (NP) experiments aim to uncover the fundamental building blocks of matter and improve our understanding of the universe. 
Many of these experiments involve high-energy collisions that can produce multiple charged particles in the presence of a magnetic field inside a charged particle detector. 
These charged particle detectors are composed of layers that record what we refer to as hits that register ionization energy deposits within the layers. 
These hits represent the coordinates at which the particles pass through the detector. 

% Overview of track reconstruction
Reconstructing the charged particle tracks constitutes an important part in the analysis of high-energy physics experiments. 
On a high level, track reconstruction is generally divided into two subtasks, namely track finding and track fitting. 
In the track finding stage, hits are grouped into subsets, each deemed to belong to one of the particles forming a track. 
In simple terms, this subtask can be viewed as connecting the right dots to each other. 
For each of the subsets provided by the track finder, the track fitting algorithm estimates a set of parameters that uniquely describe the state of the particle. 

% More on track finding in particular
Track finding helps in understanding the propagation patterns of charged particles subjected to a strong magnetic field. 
The charged particle curves in the presence of a magnetic field inversely proportional to their momentum (which is unknown at this stage); this makes it challenging, as different particles curve by different proportions making helix movement in the 3-dimensional space. 
In addition, the tracking data has noise due to background and secondary particles that may cross the particle trajectories confusing the algorithm. 

% Limitation of existing approaches
The existing algorithms used for track finding are combinatorial in nature, which do not scale with the increase in multiplicity, slowing down the entire reconstruction pipeline. 
The presence of particles with similar trajectories in addition to noise in the measurements motivates the exploration of new methods to improve performance.

% Scope of this study
In this study, we compare the performance of a Graph Neural Network (GNN) model on track finding with a traditional method used at GlueX on Forward Drift Chamber (FDC) data. 
The GlueX spectrometer~\cite{ADHIKARI2021164807} is composed of a large solenoid magnet that houses the FDC detectors. 
We demonstrate that the GNN-based track finding approach provides a 7.5\% improvement in segment efficiency at a fixed purity value compared to the traditional method while maintaining a significantly lower inference time (a 71\% reduction) by processing multiple events in batches, thereby leveraging the parallel computational capability of modern Graphical Processing Units (GPUs). 
In addition, we present the deployment of the models on a Field-Programmable Gate Array (FPGA) and demonstrate the additional speedups that can be achieved, albeit with a slight reduction in performance. 

% Paper organization
The rest of the paper is structured as follows: 
In section~\ref{sec:Previous-Work}, we describe some of the recent relevant studies found in the literature. 
Section~\ref{sec:Background} presents the detector geometry and describes the working mechanism of the traditional track finding method. 
In section~\ref{sec:Pre-Processing}, we describe our data pre-processing approach that produces graphs for GNN training and evaluation. 
Section~\ref{sec:Modeling} introduces our modeling approach and section~\ref{sec:Results} presents the performance and timing results of the proposed approach on both GPU and FPGA devices and compares them to the traditional method. Finally, we conclude with future outlook in section~\ref{sec:Conclusion}.

%% file: Previous-Work.tex
\section{Previous Work}
\label{sec:Previous-Work}

Due to the helical propagation pattern of particles in a solenoidal magnetic field, the hit projections of a given particle track on the $xy$-plane would roughly form a circle passing through the origin of the form $(x - a)^2 + (y - b)^2 = R^2 = a^2 + b^2$. 
Conformal transformation was early adopted~\cite{HANSROUL1988498} to transform circular $xy$-plane projections into linear $uv$-plane projections of the form $2au + 2bv = 1$ where $u = x / (x^2 + y^2)$ and $v = y / (x^2 + y^2)$. 
The straight lines are then used for pattern recognition to group points belonging to a given line. 
The strong assumption that the circle passes through the origin--imposed by equating $R^2$ to $a^2 + b^2$--is remedied by allowing a small difference between the two terms, which corresponds to a parabola fit with a small curvature in the $uv$-plane. 
Multiple particle scattering can be another source of deviation from the circular path in the $xy$-plane. 
The Cellular Automaton (CA) algorithm~\cite{KISEL200685} used for track finding in conformal tracking consists of building and extending cells (defined as segments connecting two hits). 
In~\cite{BRONDOLIN2020163304}, the two stages (building and extension) run recursively as the final track finding strategy. 
Another used approach is the Hough transform, where each hit corresponds to a plane in the parameter space (Hough space). 
Hits are grouped on the basis of their intersection in the Hough space. 
In practice, planes in the Hough space do not intersect at a single point. 
Therefore, the space is divided into bins and points are grouped together if their planes cross the same bin~\cite{ZHOU2025170357}. 
An overview of other traditional methods such as artificial retina and Legendre transform can be found in~\cite{Frühwirth2021}. 

Recently, Machine Learning (ML) has been introduced to address the track finding task. 
Recurrent Neural Networks (RNNs), a class of ML architectures designed for time series prediction, was adopted in~\cite{farrell2018novel} for track finding where each track is modeled as a sequence of hits. 
The task is thus formulated as a regression problem in which the RNN iteratively attempts to estimate the coordinates of the next hit given the current and previous hit coordinates. 
This method yields poor first guesses due to its inability to estimate the track trajectories. 
The detector data naturally imposes itself as a graph structure with the event hits represented as graph nodes. 
The coordinates of each hit are represented as node features in the graph. 
To this end, the same study introduced GNNs to tackle the problem. 
GNNs work by iteratively passing messages across neighboring nodes in the graph. 
Two GNN flavors were introduced. 
The \emph{first} flavor addresses the track finding task as a node classification problem. 
For a given training target track, four surrounding hits on the adjacent layer are connected. 
Seven graph iterations were executed in the GNN before a sigmoid activation was applied for each node to detect whether it belongs to the target track. 
Despite its novelty, this method requires a partially labeled graph (i.e., seeds) which limits its applicability. 
A more natural approach was adopted in the \emph{second} GNN flavor that attempts to classify edges built with a predefined set of geometric constraints between adjacent detector layers. 
Hence, only the edges connecting hits from the same particle would be predicted to be true by the GNN. 

Other studies have also demonstrated the suitability of edge-classifiers for particle tracking applications~\cite{ju2020graph, ju2021performance}. 
In~\cite{dezoort2021charged}, the same approach was followed and the constraints with which graphs were built were extensively studied. 
Previous studies implemented similar GNN architectures on FPGAs, which require special management—such as subdividing the event graph into multiple sectors—due to memory limitations \cite{heintz2020accelerated, 10.3389/fdata.2022.828666}.

%% file: Background.tex
\section{Background}
\label{sec:Background}

\begin{figure}[h!]
  \centering
  \includegraphics[width=0.8\textwidth]{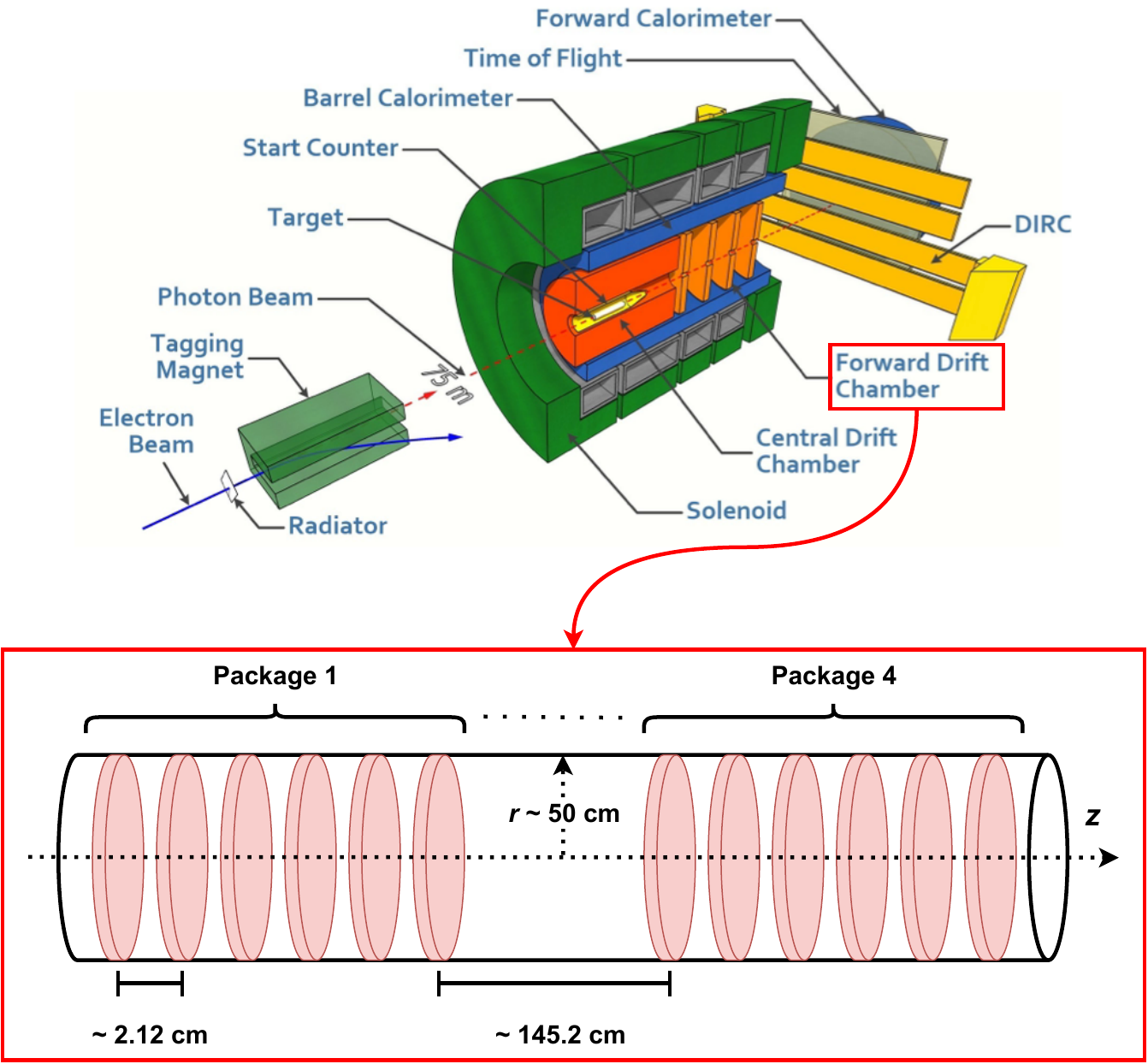}
  \caption{
  A 3-D figure illustrating the geometry of the GlueX detector. 
  It shows Forward Drift Chamber (FDC) that is composed of four packages, each including six disc-shaped layers. 
  Each of the layers has detector wires on which the ($x$, $y$, $z$) coordinates of the hits are recorded through ionization energy deposits. 
  Particles propagate along the positive $z$-axis that is perpendicular to the layers.
  }
  \label{fig:gluex_geometry}
\end{figure}

In this study, we train an edge-classifying GNN on simulated data for the GlueX detector. 
The simulation of the detector response is based on the GEANT4 software package~\cite{ALLISON2016186}.
Fig.~\ref{fig:gluex_geometry} illustrates the geometrical structure of the GlueX detector. 
A beam of high-energy photons impinges on a liquid hydrogen target in the bore of a solenoid magnet enclosing detectors for charged and neutral particle detection that form the core of the GlueX detector. 
After interactions of the beam with the target, many charged particles moving in the forward direction are reconstructed using hits in the FDC detectors. 
Hits in the FDC are separated into four packages, each containing six wire planes. 

\subsection{Traditional Method}

The traditional method used in the GlueX data analysis for track finding starts by looking for track segments in each of the four FDC packages. 
Starting with the most upstream plane in a given package containing hits, the algorithm proceeds from plane to plane looking for nearest neighbors subject to a 2 cm radial proximity threshold. 
If three or more hits are associated together, a preliminary helical fit is performed with the assumption that the particles emerged from the center of the target. 
These associated hits and the fitted helical parameters form the track segments. 
The results of the fits are used to project from one package to the next starting from the most upstream package containing track segments. 
The track segments that belong to a common track lie close to a circle in a plane perpendicular to the beam line. 
If the projected position is within a certain distance from the position of the segment in the projected package, the segments are linked together. 
The match threshold depends on the radius $r_c$ of each circle and the separation distance $d$ according to $d^2<1000/r_c$ subject to a minimum threshold value of 5 cm$^2$ and a maximum threshold value of 25 cm$^2$. 
If this match criterion is not met, the algorithm attempts to match the centers of the circles $(x_{c1},y_{c1})$ and $(x_{c2},y_{c2})$ of the two segments. 
The requirement is:
\begin{equation}
(x_{c1}-x_{c2})^2 + (y_{c1}-y_{c2})^2 < 25 \mbox{ cm}^2.
\end{equation}
Sometimes, this simple approach does not successfully link all the segments on a given track together. 
If segments in two adjacent packages are linked together and there are unmatched isolated segments or unmatched linked pairs of segments remaining, the combined set of hits are used to redo the helical fit and the new fitted results are used to project to segments downstream of the second package in the pair and the match criteria are applied again. 
Each set of linked segments forms a track candidate.

\subsection{ML Pipeline Overview}
\begin{figure}[h!]
  \centering
  \includegraphics[width=1.0\textwidth]{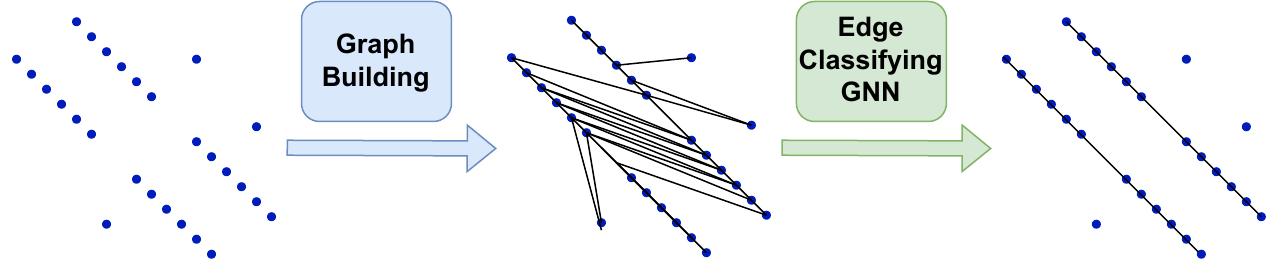}
  \caption{ML Pipeline Summary: The detector yields hit coordinates information. The graph-builder constructs graph from the raw hits by introducing edges between some of the hits based on graph building constraints. Finally, the edge-classifier GNN should discard false edges only (i.e., edges not corresponding to track segments).}
  \label{fig:ML_pipeline}
\end{figure}

The simple approach described in the previous subsection provides a reasonable level of efficiency and purity in track finding but there is still scope to improve the performance with ML methods such as GNN. 
Fig.~\ref{fig:ML_pipeline} presents an overview of the pipeline workflow presented in this paper. 
The details of the two main components of the pipeline, namely the graph-builder and the GNN model, are discussed in the following two sections. 

%% file: Pre-Processing.tex
\section{Data Pre-processing (Graph Building)}
\label{sec:Pre-Processing}

The data acquired from GlueX FDC (depicted in Fig.~\ref{fig:gluex_geometry}) for each event is presented as a collection of hits. 
Each hit has 3-dimensional spatial coordinates ($x$, $y$, $z$) at which the hit is recorded. 
In the simulation data, the identity of the particle associated with each hit is known, which enables assigning ground-truth label for each edge. 
An edge is given the label "true" if both associated hits it connects belong to the same particle (i.e., constituting a particle track segment), and "false" otherwise. 
As such, a graph can be constructed by creating edges between hits on adjacent layers. 

Creating edges between all possible adjacent hits would cause a high imbalance between true and false edges in addition to demanding higher computational resources to process these dense graphs. 
The solution is to create an edge connecting two hits, $a$ and $b$, only if it meets constraints based on the detector geometry. 
For our case, the constraints are as follows:
\begin{itemize}
    \item $d^{(a,b)}_{xy} < 34.4$ cm, where $d^{(a,b)}_{xy}$ is the Euclidean distance between the two hits in the $xy$-plane. 
    The $z$ dimension is not included due to the non-uniformity of the distance between the detector layers as shown in Fig.~\ref{fig:gluex_geometry}. 
    Layers belonging to different groups have much larger separation compared to layers within the same group. 
    Additionally, the inclusion of skip edges that will be introduced later in this section introduces non-uniformity in the $z$-distance even among layers within the same group.

    \item $d^{(a,b)}_{xy} / dz^{(a,b)} < 5.4$ where $dz^{(a,b)}$ refers to the physical distance between the layers on which the two hits are recorded. 
    While the previous constraint limits the absolute Euclidean distance in the $xy$-plane, this constraint further limits the $xy$-plane distance for the hits that are close in $z$ dimension.
    
    \item $|d\phi^{(a,b)}| < 2.3$ rad where $d\phi^{(a,b)}$ is the difference in the azimuth angle of the two hits (i.e., $\phi_b - \phi_a$).
\end{itemize}

The selection of the cut-off points for the various constraints, as listed above, is based on an empirical analysis with a multi-objective genetic algorithm~\cite{MOGA} to optimize both efficiency and purity on realistic simulation data. 
The details of this study are described in Appendix~A. 
Each built event graph, $g$, has node features $g.X \in \mathbb{R}^{|\mbox{hits}| \times 3}$ where 3 refers to the used cylindrical coordinates ($r$, $\phi$, $z$), and adjacency list of edges $g.E \in \mathbb{R}^{|\mbox{edges}| \times 2}$. 
$|\mbox{hits}|$ and $|\mbox{edges}|$ refer to the cardinality of the sets of event hits and edges, respectively. 
If $g$ is constructed from a simulated event, it would additionally have $g.label \in \mathbb{R}^{|\mbox{edges}|}$.

\begin{figure}[h!]
\centering

\begin{subfigure}{\textwidth}
\centering
  \includegraphics[width=1.0\textwidth]{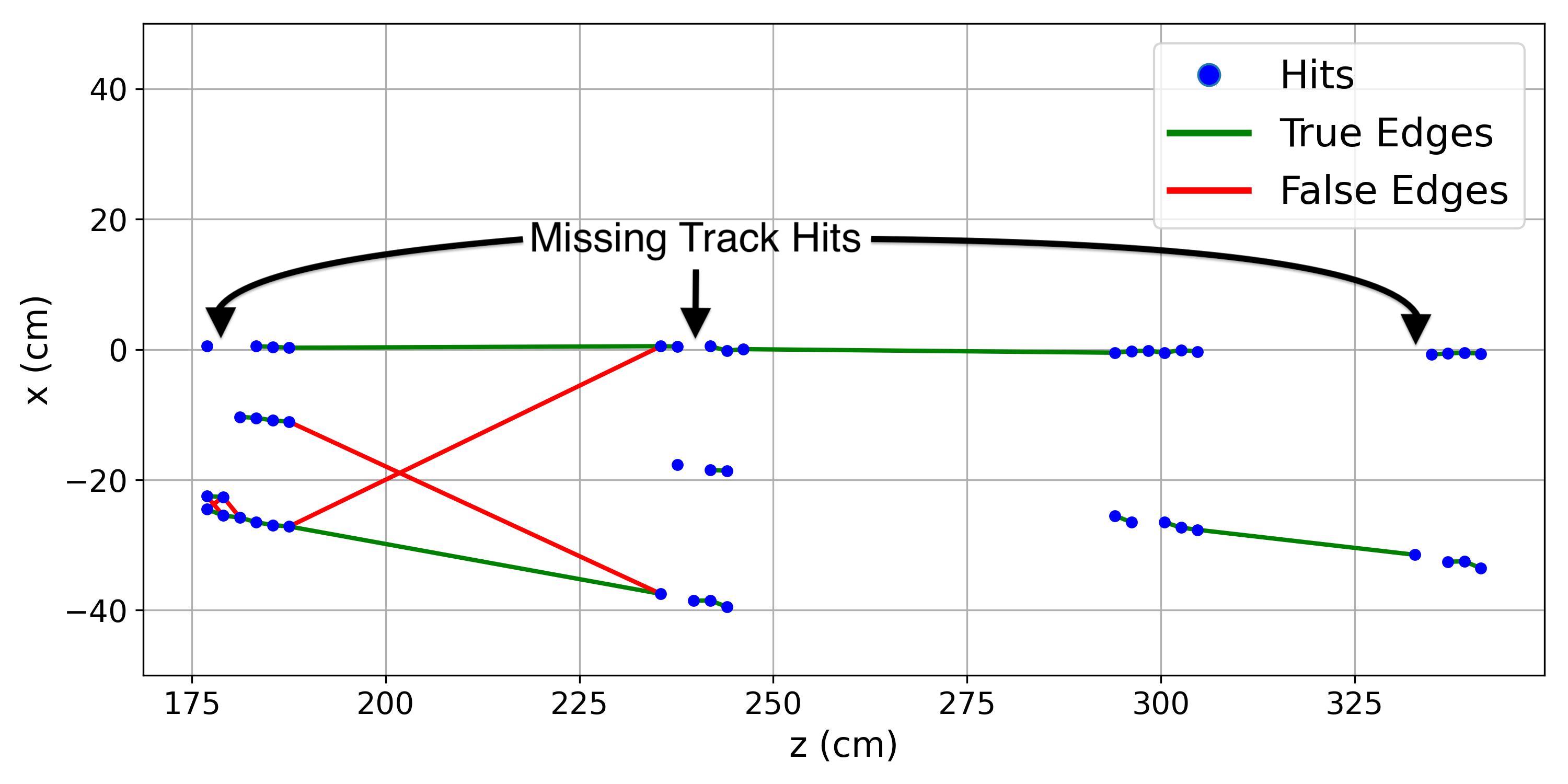}
  \subcaption{Event graph built without skip edges (i.e., $skip_{max} = 0$)}
  \label{sub-fig:skip_max_0}
\end{subfigure}

\begin{subfigure}{\textwidth}
\centering
  \includegraphics[width=1.0\textwidth]{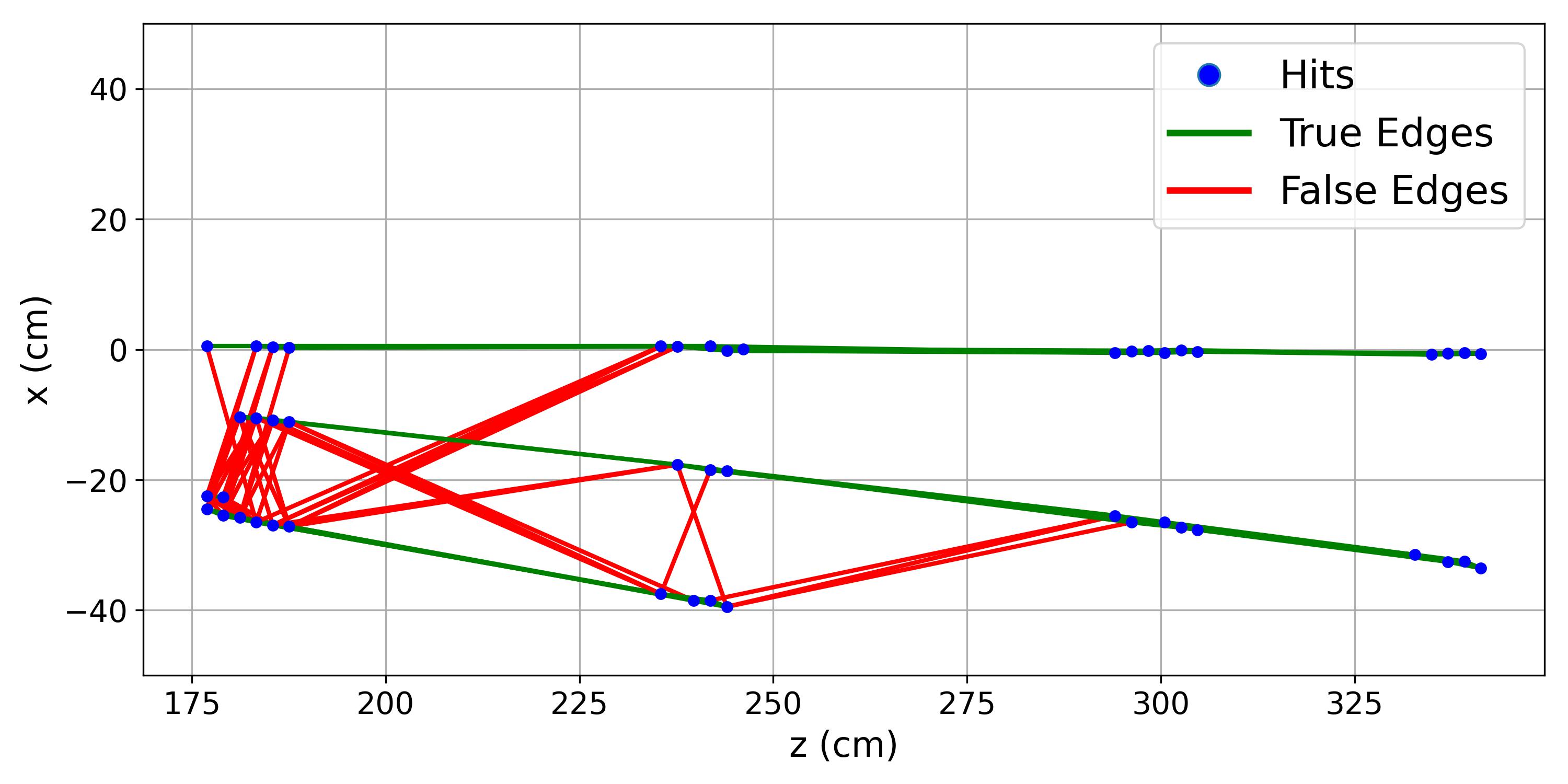}
  \subcaption{Same event graph built with $skip_{max} = 2$}
  \label{sub-fig:skip_max_2}
\end{subfigure}

\caption{Graphs built with no skip edges (i.e., $skip_{max} = 0$) are more susceptible to the problem of missing hits as shown in part (a). This problem is resolved by introducing edges between hits on non-consecutive layers with a maximum separation of $1 + skip_{max}$. Part (b) shows the same event built with a $skip_{max}$ of 2 that does not suffer from the gap patterns which makes detecting the full track possible (i.e., higher efficiency).}
\label{fig:event_25_simulated_event_graphs}
\end{figure}

In GlueX, limited detector efficiency causes some tracks to have missing hits. 
This causes gap patterns along those tracks as demonstrated in Fig.~\ref{sub-fig:skip_max_0} where the simulated event graph is built by considering only edges connecting hits on consecutive layers. 
Note that the role of the GNN is limited to classifying existing edges in the input graph (i.e., no additional edges are introduced). 
To resolve this problem, we introduce the notion of skip edges (or residual edges) that allows connecting hits on non-consecutive detector layers. 
$skip_{max}$ is a parameter that defines the maximum number of intermediate layers an edge can cross (i.e., $0 < layer_b - layer_a \leq 1 + skip_{max}$ where $layer_i$ refers to the order of the detector layer on which hit $i$ is located). 
Fig.~\ref{sub-fig:skip_max_2} shows the same event as Fig.~\ref{sub-fig:skip_max_0} but built with $skip_{max}$ value of 2. 
This comes with the cost of building more dense graphs in addition to introducing redundant long edges by connecting two hits directly with an edge that are already connected with multiple shorter edges. 
This can be addressed in a post-processing step after the ML algorithm rejects the false edges.

%% file: Modeling.tex
\section{Edge-Classifier GNN}
\label{sec:Modeling}

\begin{figure}[h!]
    \centering

    % Row 1: One large subfigure
    \begin{subfigure}[t]{0.8\textwidth}
        \centering
        \includegraphics[width=\linewidth]{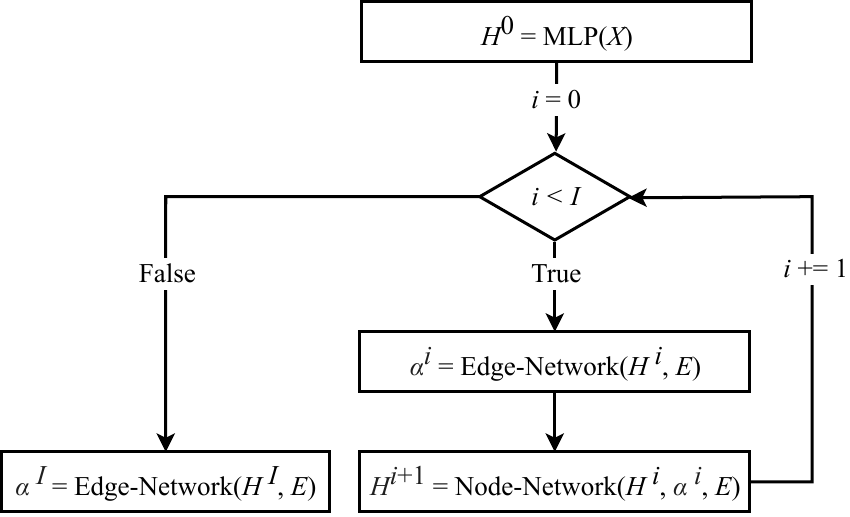}
        \caption{Edge-Classifier$(X, E, I) \rightarrow \alpha^I$}
        \label{sub-fig:edge_classifier}
    \end{subfigure}
    
    \vspace{1em} % Adjust space between rows if needed
    
    %% Row 2: Two small subfigures
    \begin{subfigure}[t]{0.45\textwidth}
        \centering
        \includegraphics[width=\linewidth]{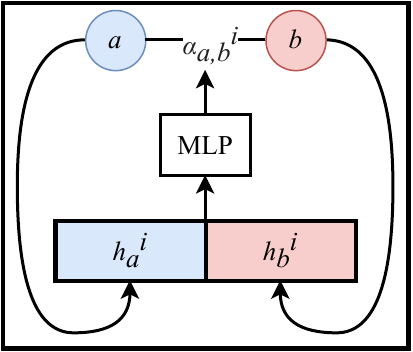}
        \caption{Edge-Network$(H^i, E) \rightarrow \alpha^i$}
        \label{sub-fig:edge_network}
    \end{subfigure}
    %\hfill
    \begin{subfigure}[t]{0.45\textwidth}
        \centering
        \includegraphics[width=\linewidth]{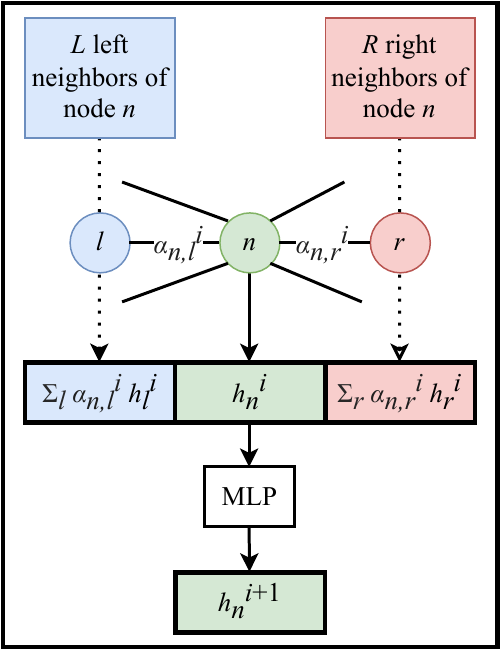}
        \caption{Node-Network$(H^i, \alpha^i, E) \rightarrow H^{i + 1}$}
        \label{sub-fig:node_network}
    \end{subfigure}

    \caption{Edge-Classifier GNN takes a graph (with $X$ and $E$ representing coordinates of graph hits and adjacency list, respectively) and returns the probability of each edge being true after performing $I$ message passing iterations.}
    \label{fig:block_diagram}
\end{figure}

The edge-classifier GNN architecture depicted in Fig.~\ref{sub-fig:edge_classifier} takes the built event graphs as input and is trained in a supervised fashion to output the correct label for each edge. 
The classifier starts by expanding the dimensionality of the original hit features (i.e., cylindrical coordinates of each hit) and then alternates between edge-network (Fig.~\ref{sub-fig:edge_network}) and node-network (Fig.~\ref{sub-fig:node_network}) $I$ times, where $I$ represents the number of message passing iterations carried out by the GNN. 

For each edge in the adjacency list, $E$, the edge-network takes concatenated embeddings of the two associated hits, $a$ and $b$, and assigns a weight to that edge in the [0-1] range, denoted by $\alpha_{a,b}^i$ where $i$ is the index of the message passing iteration. 
Hence, a sigmoid activation function is used at the end of the multilayer perceptron (MLP) of the edge-network. 
At the end of the edge-classifier, this weight represents the probability of this edge being true. 
The $\alpha$ weights are also used to scale the messages passed between the nodes in the node-network that updates the embeddings of all hit nodes in every iteration. 
For a particular hit $n$, three embeddings are concatenated: 
\emph{\textbf{i)}} Aggregated scaled embeddings from left neighbor hits located on layers $< layer_n$, 
\emph{\textbf{ii)}} Embedding of hit $n$ itself, and 
\emph{\textbf{iii)}} Aggregated scaled embeddings from right neighbor hits on layers $> layer_n$. 
The concatenation of those three embeddings is then forwarded to the MLP of the node-network. 
The aggregation of the embeddings is performed via a gather reduction operation. 

Before applying the edge-network, we concatenate the current node representation $H$ with the original node features $X$ (i.e.\ \(H \leftarrow [H;X]\)) to preserve the initial feature information. 
Table~\ref{tab:model_parameters} presents the different parameters used for training the model.

\begin{table}[ht]
    \centering
    \begin{tabular}{l || c}
        Model Parameter & Value\\
        \hline
        \hline
        Number of Message Passing Iterations ($I$) & 1 \\
        MLP Hidden Layers Width ($W$) & 128 \\
        MLP Depth ($D$) & 3 \\
        Default Activation & Rectified Linear Unit (ReLU)~\cite{nair2010rectified} \\
        Edge-Network Output Activation & Sigmoid \\
        Dropout Probability & 0.05 \\
        Loss & Binary Cross-Entropy \\
        Optimizer & Adam~\cite{kingma2014adam} \\
        Learning Rate & 0.001 \\
    \end{tabular}
    \caption{Model Parameters}
    \label{tab:model_parameters}
\end{table}

%% file: Results.tex
\section{Results \& Discussion}
\label{sec:Results}

\subsection{Traditional Method vs ML Pipeline}
\label{subsection:trad_vs_pipeline}

% Traditional Method Performance
The simulated dataset consists of 90,842 events, which are divided into training, validation, and testing sets in approximate proportions of 70\%, 15\%, and 15\%, respectively. 
Each event contains lists of hits encoding positions from multiple tracks. 
We begin by evaluating the traditional method on the testing simulated events in terms of efficiency and purity. 
Efficiency is defined as the fraction of true edges retained by the system (traditional method, graph-builder, or ML pipeline) whereas purity refers to the accuracy when a prediction is made. 
The efficiency and purity presented in this study are calculated on the segment (edge) level. 
The reported efficiency and purity of the traditional method on the test dataset are 0.9119 and 0.9462, respectively. 
Unlike the ML pipeline that outputs the probability score for each edge, this method produces edges as part of predicted tracks that can be treated as binary labels. 

\begin{figure}[h!]
  \centering
  \includegraphics[width=0.75\textwidth]{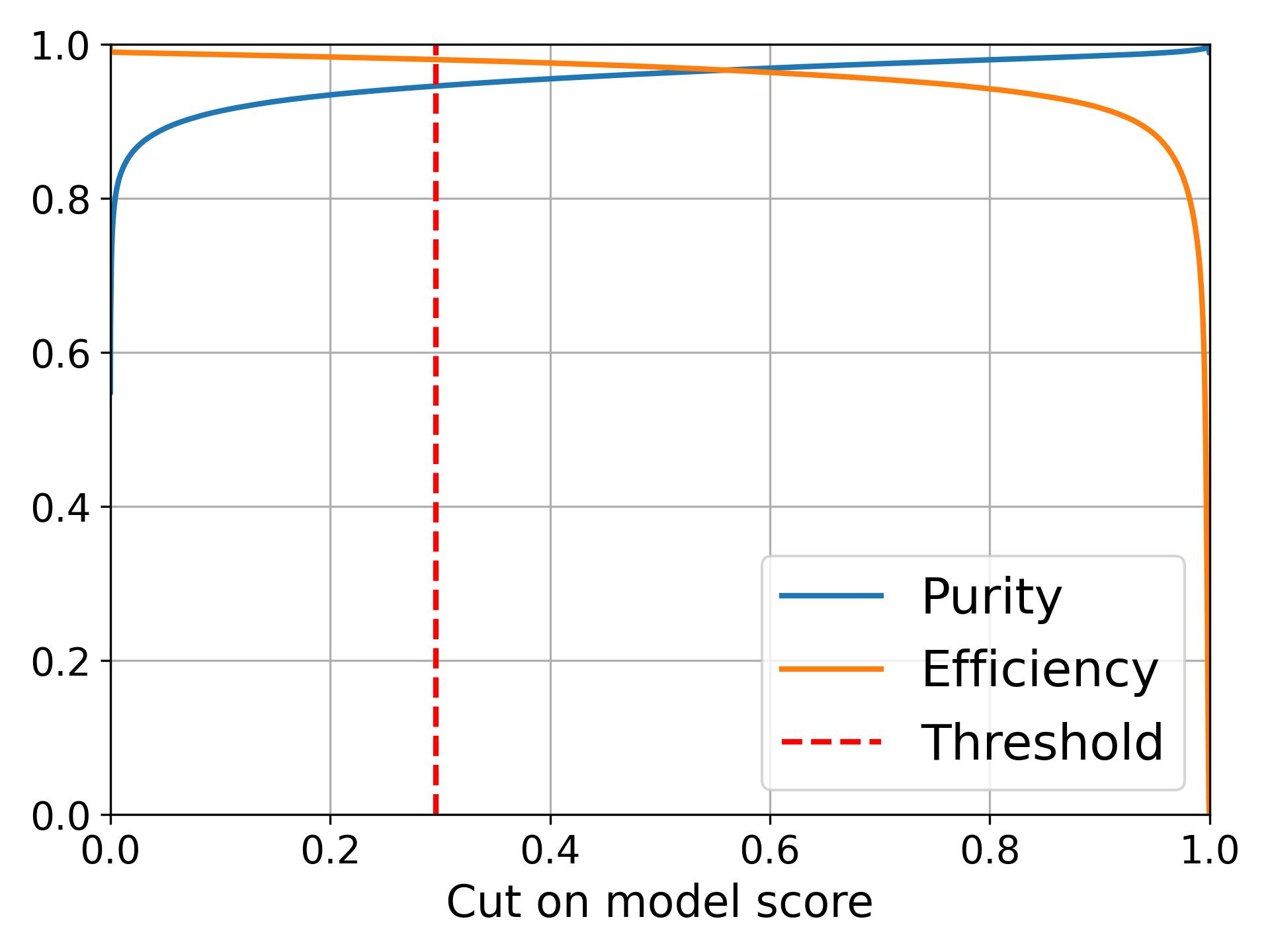}
  \caption{Overall efficiency \& purity of the ML pipeline evaluated on simulated test events. At a fixed purity level of 0.9462, the pipeline achieves efficiency of 0.9806 compared to 0.9119 scored by the traditional method (7.5\% increase) at a threshold $\approx$ 0.3.}
  \label{fig:combined_eff_purity}
\end{figure}

% Pipeline & its performance comparison to traditional method (simulated & real)
The efficiency and purity of the graph-builder with the constraints listed in section~\ref{sec:Pre-Processing} are 0.9905 and 0.5473, respectively. 
To limit the inference time of the model, shallow MLP modules with a depth, $D$, of 3 are used within the classifier GNN. 
To compensate for the reduced expressivity, we adopt a relatively large width (i.e., $W$ = 128) compared to the input dimensionality, which corresponds to the three positional coordinates. 
Fig.~\ref{fig:combined_eff_purity} shows the efficiency and purity of the pipeline (i.e., Graph-Builder + Edge-Classifier) as a function of the probability score produced by the model. 
To compare the pipeline with the traditional method, we applied a threshold on the output that yields the same purity value (0.9462). 
The corresponding efficiency is 0.9806 which represents a 7.5\% increase over the 0.9119 efficiency of the traditional method. 

\begin{figure}[h!]
  \centering
  \includegraphics[width=0.8\textwidth]{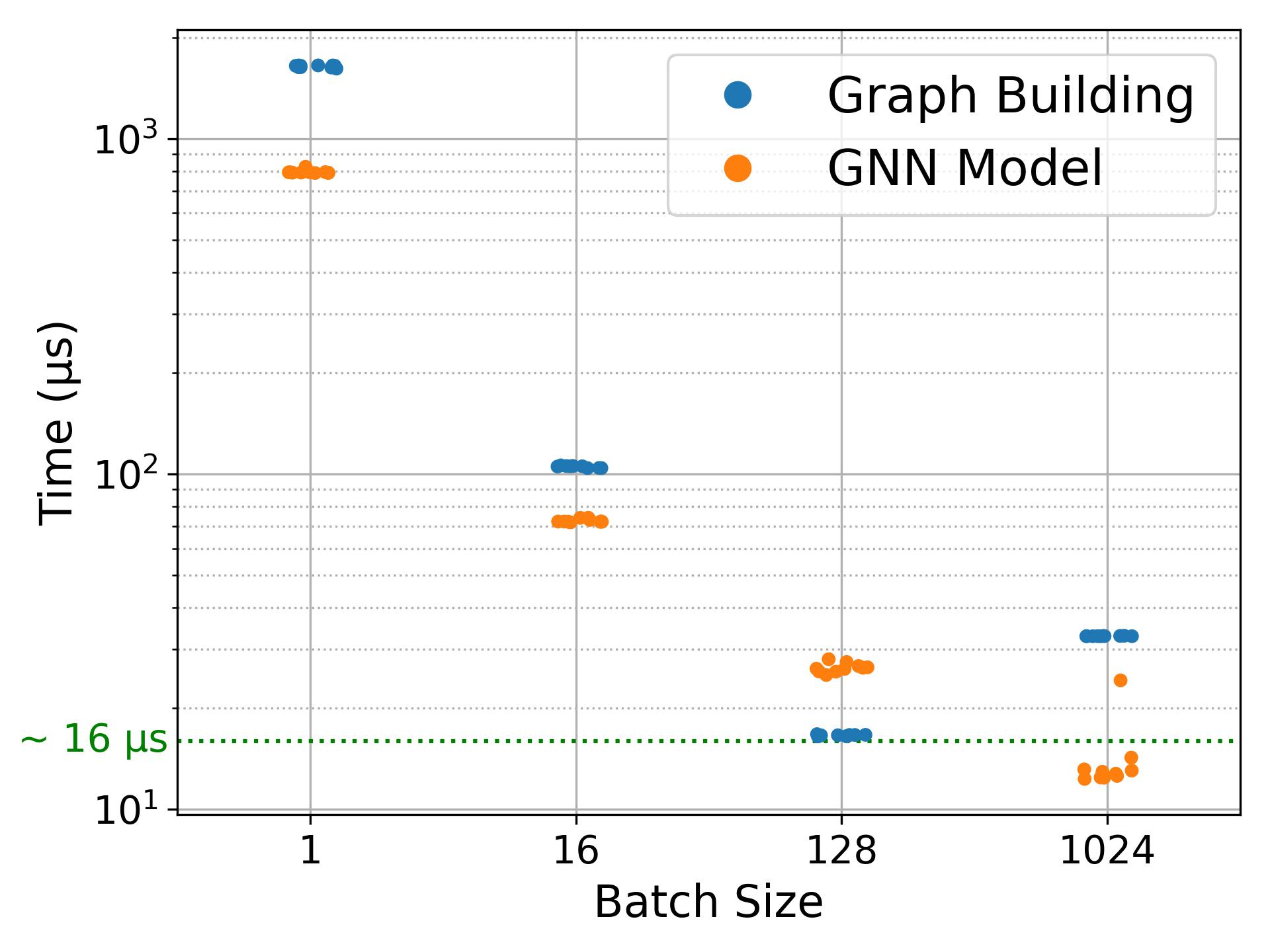}
  \caption{A100 GPU execution time per GlueX event. The analysis is conducted on 16,384 events and repeated across 10 trials, each represented as a dot in the figure. Significant speedups are observed with increasing batch size. Among the studied batch sizes, the optimal was found to be 128, yielding an overall execution time of approximately 44~$\mu$s per event. Beyond this point, the graph construction time begins to rise again, negatively impacting the total pipeline speed.}
  \label{fig:execution_time}
\end{figure}

% Speedup: Batching
To speed up the pipeline, we introduced a batched implementation of both of its phases that takes advantage of the strong parallel computational capability of modern GPUs. 
In the graph building stage, graphs of multiple events are built simultaneously by introducing an additional constraint that ensures that no edge connects hits belonging to different events. 
Once a batched graph is created by the graph-builder, it gets directly forwarded to the trained GNN model in the inference pipeline. 
Fig.~\ref{fig:execution_time} shows the execution time of the graph building and GNN model (with a single message passing iteration, i.e., $I$ = 1) taken by A100 GPU for different batch sizes. 
At a batch size of 128, the overall pipeline execution time per GlueX event is approximately 44~$\mu$s ($\pm$~1~$\mu$s), which includes data transfer to the A100 GPU, graph construction, and GNN model inference. 
This represents a significant speedup of approximately 71\% compared to the traditional method, which requires approximately 152~$\mu$s ($\pm$~1~$\mu$s) per GlueX event.

% Compare with I = 3
The demonstrated performance and timing results of the ML pipeline used a GNN model with a single message passing iteration (i.e., $I = 1$). 
Generally, the number of message passing iterations is an important hyperparameter in GNN models. 
Increasing this parameter allows each node to aggregate information from more distant nodes, thereby gaining broader awareness of its graph neighborhood. 
However, setting this number too high can lead to the common issue of over-smoothing, where node representations become indistinguishably similar across the graph~\cite{over_smoothing_1, over_smoothing_2}. 
To mitigate this effect, we concatenate the original embedding, $X$, to the node embedding after every message passing iteration as described earlier in section~\ref{sec:Modeling}. 
To investigate the effect of $I$ in our setting, we trained a similar GNN model with 3 message passing iterations and evaluated its speed and performance. 
The model inference time on A100 GPU increased by more than a factor of 2 which is expected given the inherently sequential nature of the message passing mechanism. 
Meanwhile, the yielded efficiency at the same purity level (0.9462) was 0.9857 (compared to 0.9806), representing an improvement of less than 0.6\%. 
We therefore conclude that increasing $I$ is not justified, as the marginal gain in performance comes at the cost of a significant slowdown. 
This could be attributed to the shortcut edges that are introduced by the $skip_{max}$ parameter when building the graphs, eliminating the need for extended $I$. 
For example, with a $skip_{max}$ of 3, a certain node becomes aware of up to its fourth degree neighbors after a single message passing iteration.

\begin{figure}[h!]
\centering

\begin{subfigure}{\textwidth}
\centering
  \includegraphics[width=1.0\textwidth]{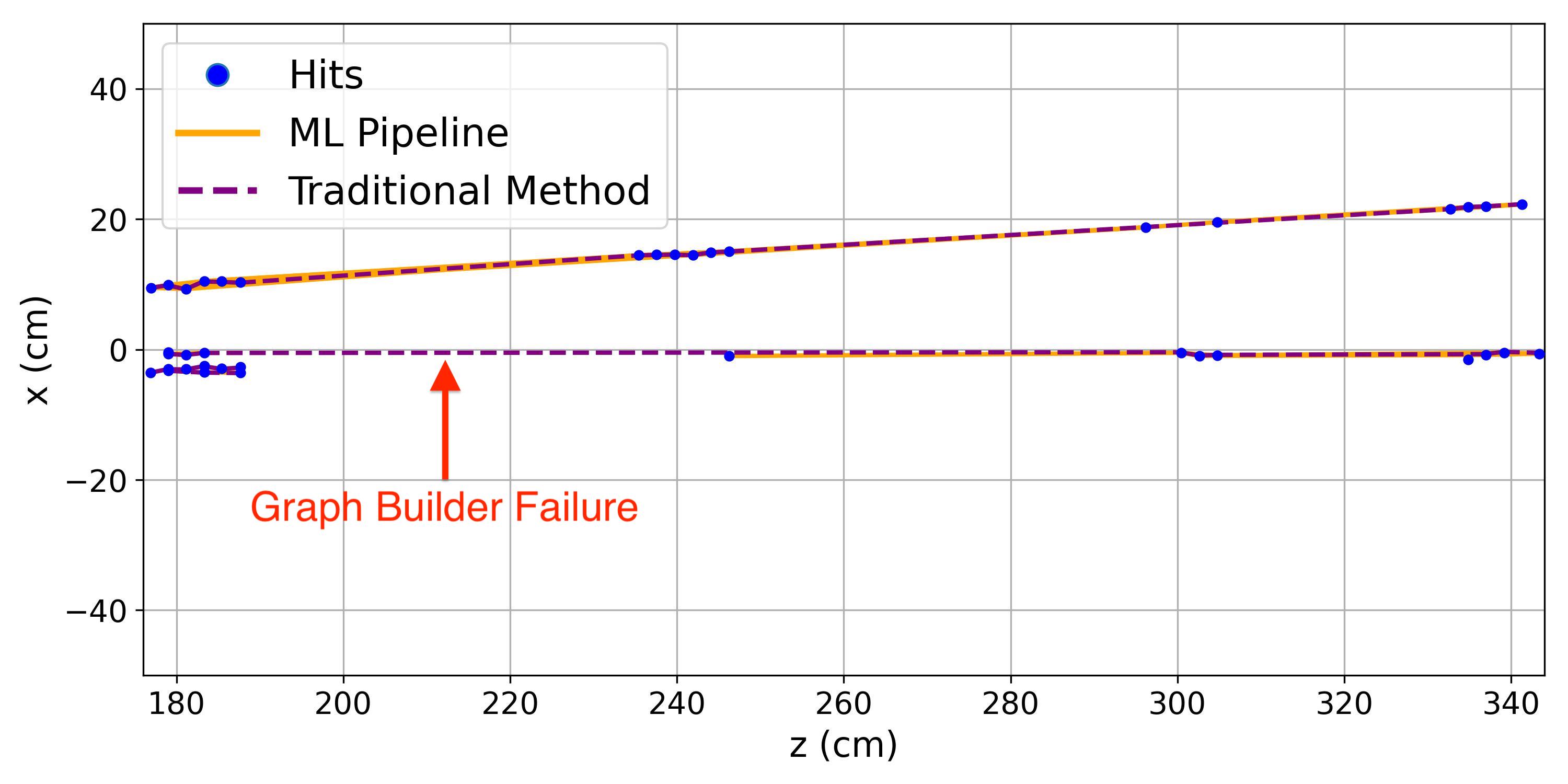}
  \subcaption{Graph-Builder with $skip_{max}$ of 3 does not create edge with 7 consecutive missing hits. ML may not connect distant hits that have number of consecutive missing hits more than $skip_{max}$ parameter.}
  \label{sub-fig:run_66_real_example_event_30}
\end{subfigure}

\begin{subfigure}{\textwidth}
\centering
  \includegraphics[width=1.0\textwidth]{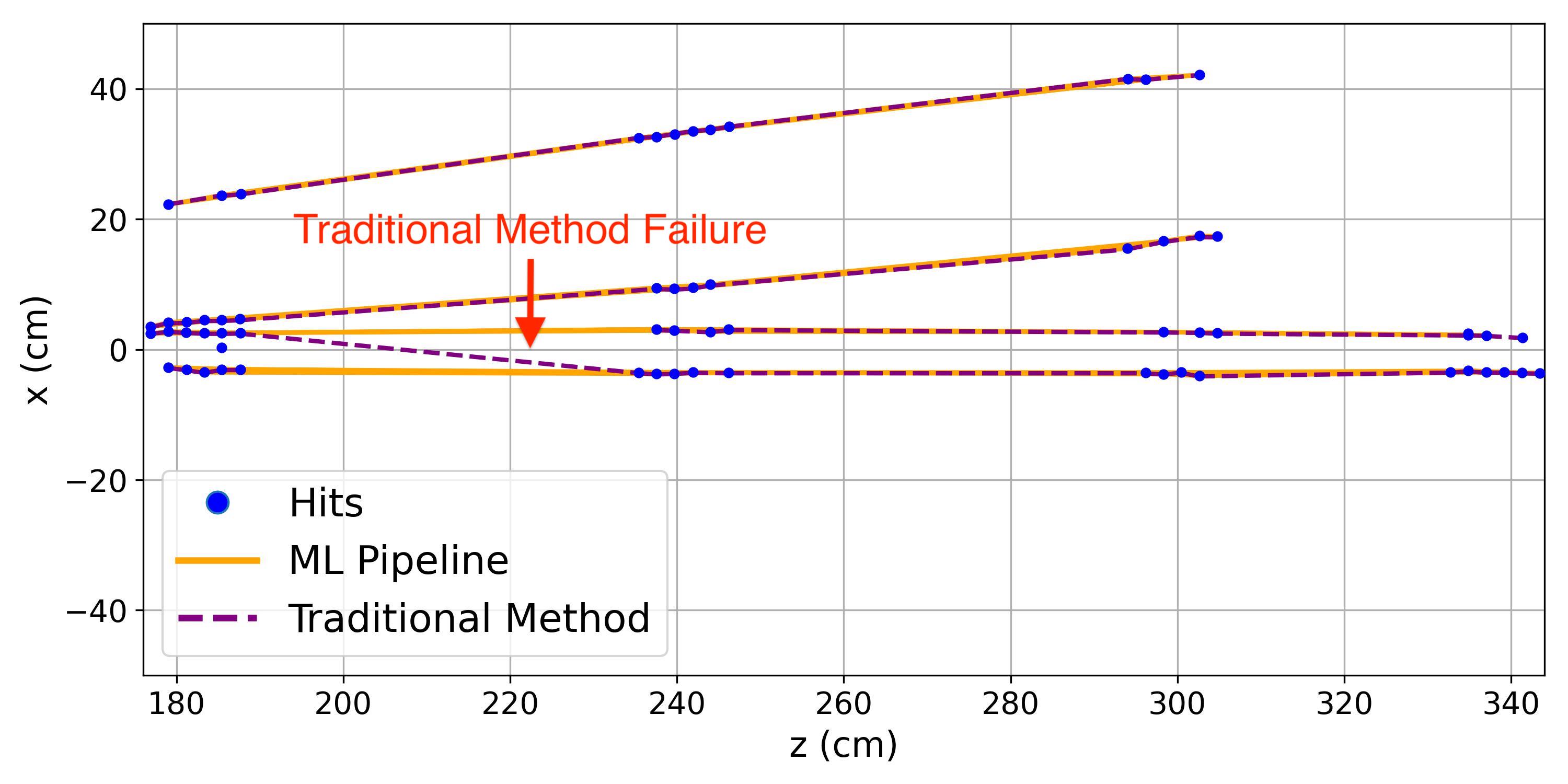}
  \subcaption{Traditional method mistakenly merging two different tracks}
  \label{sub-fig:run_66_real_example_event_24}
\end{subfigure}

\caption{Real GlueX examples}
\label{fig:real_gluex_examples}
\end{figure}

Fig.~\ref{fig:real_gluex_examples} shows two demonstrative examples of events from GlueX detector that compare the traditional method with the ML pipeline by highlighting their limitations. 
The event shown in Fig.~\ref{sub-fig:run_66_real_example_event_30} demonstrates an instance where the event graph was built with a $skip_{max}$ parameter of 3 whereas the track had 7 consecutive missing hits, as such graph-builder does not connect the distant hits. 
To fix this, it is generally desired to increase this parameter to at least the possible number of consecutive missing hits. 
This solution is only valid on the simulated data with available ground-truth information. 
It is worth reiterating that setting the $skip_{max}$ parameter to an arbitrarily large number can significantly increase the number of redundant edges, resulting in extremely dense graphs that are computationally expensive to process. 
Fig.~\ref{sub-fig:run_66_real_example_event_24} on the other hand shows an event in which the traditional method merges two distinct particle tracks. 
As a result, the two tracks were not properly captured.

\subsection{FPGA}
\label{subsection:FPGA}

The motivation for using FPGA lies in achieving even greater speed improvements compared to the A100 GPU. 
While the models presented earlier demonstrate strong performance, their large size renders them infeasible for deployment on FPGAs, which are constrained by limited hardware resources. 
To determine the largest model that could fit on our FPGA, we conducted a simple study by varying the width, $W$, of the MLP modules in the edge-classifier GNN. 
Several randomly instantiated models, each with a different $W$, were converted into FPGA-compatible C++ code using an open-source Python package named High-Level Synthesis (HLS) for ML, widely known as \texttt{hls4ml}~\cite{fastml_hls4ml}.
Each model was designed to process events with up to 150 nodes and 256 edges. 
The U200 FPGA board was used to obtain resource estimates during synthesis, whose maximum resources can be seen in Table~\ref{tab:u200MaxRes}

\begin{table}[ht]
    \centering
    \begin{tabular}{r|r|r|r|r}
        BRAM & DSP & FF & LUT & URAM \\
        \hline
        \hline
        4320 & 6840 & 2364480 & 1182240 & 960
    \end{tabular}
    \caption{Maximum resource values of the U200 part ``xcu200-fsgd2104-2-e'' used for synthesis}
    \label{tab:u200MaxRes}
\end{table}

\begin{figure}[!h]
    \centering
    \includegraphics[width=0.8\textwidth]{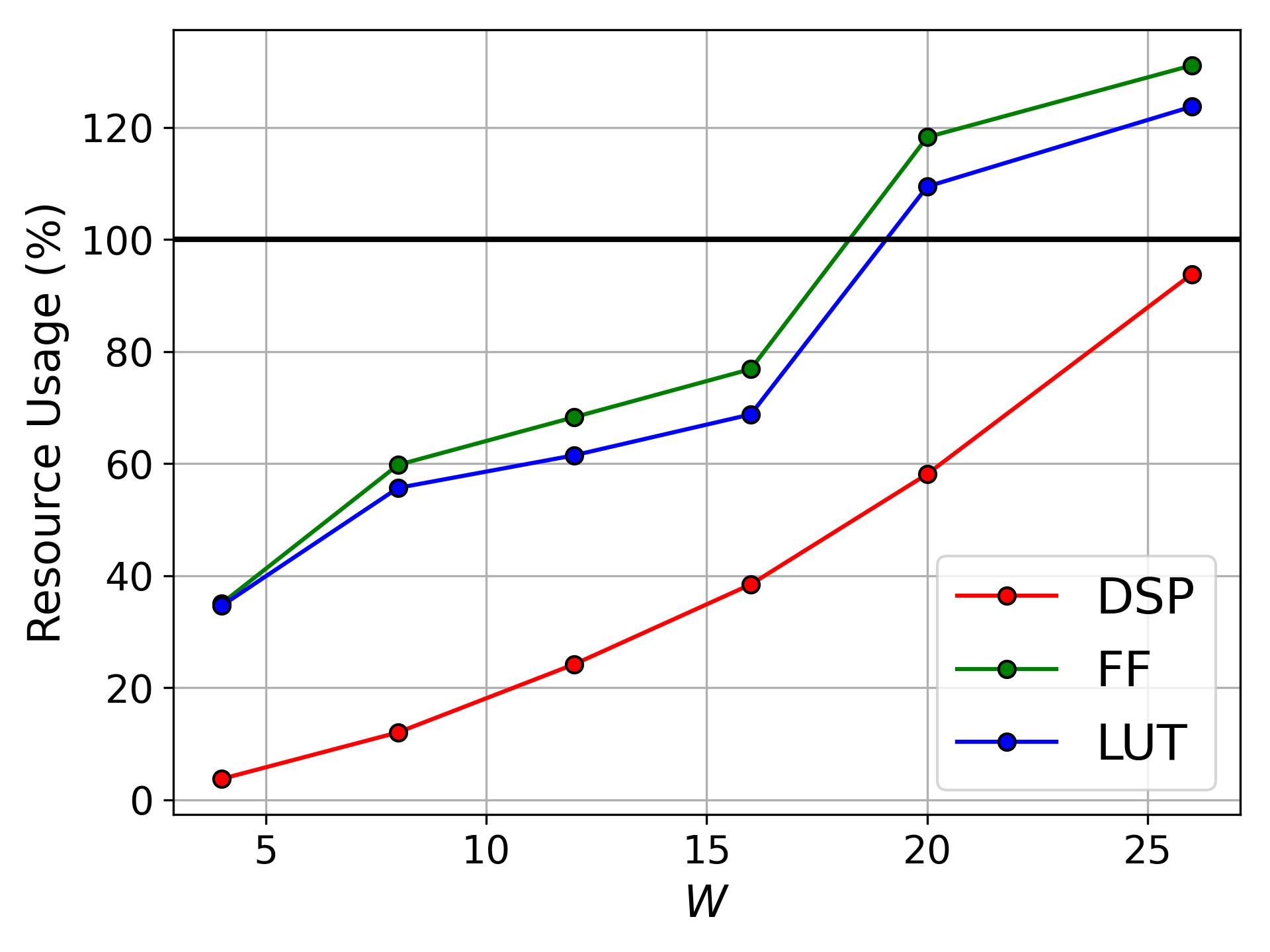}
    \caption{Digital Signal Processing (DSP), Flip-Flop (FF), and Look Up Table (LUT) resources usage on FPGA. $W$ represents the width of the hidden layers of MLP modules in the GNN model. At $W = 16$, FF and LUT resources become nearly saturated.}
    \label{fig:fpga_resources}
\end{figure}

Fig.~\ref{fig:fpga_resources} shows the different resource usage estimates for different $W$ values, namely, Digital Signal Processing (DSP), Flip-Flop (FF), and Look Up Table (LUT) when Dataflow is enabled, an optimization technique that allows concurrent execution of high-level functions, significantly improving throughput and reducing Initiation Interval (II). 
The estimates were obtained using Vitis HLS tool~\cite{amd_vitis_hls}. 
For FPGA deployment, we chose a pre-trained model with $W = 16$ where the FF and LUT resources become nearly saturated. 
Table~\ref{tab:resourcesFPGA} shows that the model utilizes up to 77\% of the memory resources (i.e., FF) while significantly reducing the time between successive outputs (i.e., II) to 2.5~$\mu$s.  
This represents a substantial speedup compared to the A100 GPU inference time, as shown in Fig.~\ref{fig:execution_time}). 
However, this speedup comes at the cost of a slight reduction in efficiency (0.9565 vs. 0.9806) at the same purity level of 0.9462, attributed to the reduced model capacity. 
Furthermore, we have not fully developed and optimized graph building for FPGA yet, as such these plots only show the model resource usage and timing performance.

\begin{table}[ht]
    \centering
    \begin{tabular}{l || c | c | c | c | c}
         & \shortstack{Latency \\ cycles; $\mu$s} & \shortstack{II \\ cycles; $\mu$s} & \shortstack{\% DSP \\ \strut} & \shortstack{\% FF \\ \strut} & \shortstack{\% LUT \\ \strut}\\
        \hline
        \hline
        \texttt{edge\_classifier} & 1713; 8.57 & 503; 2.52 & 38.44 & 76.87 & 68.78\\
        \hline
        \texttt{READ\_X} & 452; 2.26 & 452; 2.26 & 0.00 & 0.80 & 0.35\\
        \texttt{READ\_E} & 502; 2.51 & 502; 2.51 & 0.00 & 0.34 & 0.40\\
        \texttt{edge\_classifier\_mlp} & 160; 0.80 & 160; 0.80 & 6.39 & 4.73 & 1.99\\
        \texttt{edge\_network\_i} & 417; 2.09 & 417; 2.09 & 9.33 & 18.71 & 11.40\\
        \texttt{node\_network} & 161; 0.81 & 161; 0.81 & 13.95 & 11.24 & 8.01\\
        \texttt{edge\_network\_I} & 265; 1.33 & 265; 1.33 & 8.77 & 5.92 & 7.03\\
        \texttt{WRITE\_prediction} & 252; 1.26 & 252; 1.26 & 0.00 & 0.00 & 0.12\\
    \end{tabular}
    \caption{
    Resources and timings of various functions on the FPGA with the same hyperparameters specified previously ($I = 1$, $W = 16$, $D = 3$). 
    \texttt{READ\_X} and \texttt{READ\_E} refer to reading node features and adjacency list from the data stream, respectively. 
    \texttt{WRITE\_prediction} refers to writing last edge-network call (i.e., \texttt{edge\_network\_I}) predictions to the data stream. 
    }
    \label{tab:resourcesFPGA}
\end{table}

%% file: Conclusion.tex
\section{Conclusion \& Future Work}
\label{sec:Conclusion}

In this study, we presented a Machine Learning (ML) pipeline to solve the challenging track finding problem. 
The pipeline is composed of: 1) Graph-Builder that addresses the missing track hits problem resulting from the limited efficiency of the detector by introducing skip edges, and 2) Edge-Classifier Graph Neural Network (GNN) that learns to filter out false edges. 
By batching multiple event graphs, the parallel computational resources of modern GPUs are leveraged, resulting in a significant reduction of the average ML pipeline inference time compared to the traditional method used for track finding in the GlueX experiment at Jefferson Lab (44~$\mu$s vs 152~$\mu$s). 
On simulated data, the pipeline outperforms the traditional method by yielding a 7.5\% higher efficiency at a fixed purity level. 
Based on expert's opinion on some GlueX examples, the pipeline is less susceptible to merging distinct tracks. 
In addition, FPGA implementation was explored to reduce the inference time even further at a slight cost of reduced efficiency compared to the big model used on GPU. 

In the future, we will explore deploying the pipeline in the experimental halls using both GPUs and FPGAs. 
GNN model will be updated to include uncertainty quantification with each prediction. 
We also aim to address the track fitting problem that accounts for the majority of reconstruction time via novel ML architectures. 
Finally, we will apply this method on high track multiplicity experiments where significant speed improvements are expected compared to the combinatorial traditional method.

%% file: Acknowledgements.tex
\section{Acknowledgments}
This manuscript has been authored by Jefferson Science Associates (JSA) operating the Thomas Jefferson National Accelerator Facility for the U.S. Department of Energy under Contract No. DE-AC05-06OR23177. 
The authors acknowledge support by the U.S. Department of Energy, Office of Science. 
The US government retains and the publisher, by accepting the article for publication, acknowledges that the US government retains a nonexclusive, paid-up, irrevocable, worldwide license to publish or reproduce the published form of this manuscript, or allow others to do so, for US government purposes. 
DOE will provide public access to these results of federally sponsored research in accordance with the DOE Public Access Plan (http://energy.gov/downloads/doe-public-access-plan).

%% file: Appendix.tex
\appendix
\label{app:MOGA}

\section{Graph-Builder Constraint optimization using genetic algorithm}

\begin{figure}[h]
    \centering
    \includegraphics[width=0.8\linewidth]{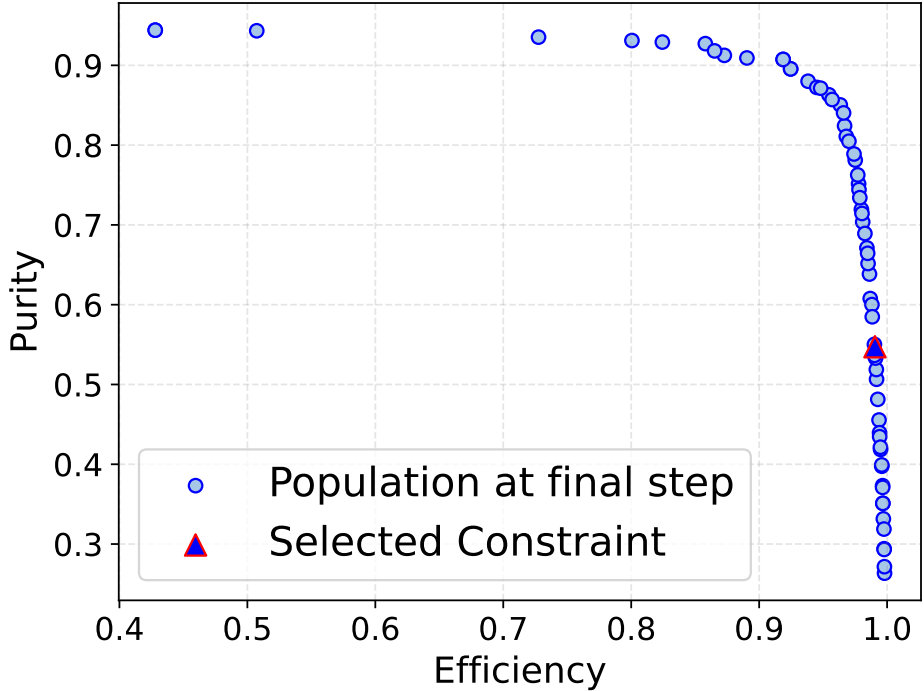}
    \caption{Pareto front of optimal efficiency and purity produced at the final step of MOGA optimization over the graph-builder constraints}
    \label{fig:pareto}
\end{figure}

The determination of graph-builder constraint values is driven by the need to optimize efficiency and purity. 
The role of the edge-classifier GNN is limited to filtering out the false edges without introducing any new edges. 
As such, the input graphs should have maximum efficiency. 
Graph builder can reduce some of the unwanted edges by applying some constraints based on detector geometry. 
However, there is a trade-off between efficiency and purity. 
This competing relationship is depicted in Fig.~\ref{fig:pareto}, which presents the Pareto front generated using the NSGA-II~\cite{NSGA-II} Multi-Objective Genetic Algorithm (MOGA)~\cite{MOGA}.

Multi-Objective Optimization (MOO) problems are characterized by solutions that form a Pareto front. 
This front consists of solutions where improving one objective necessarily compromises another. 
Essentially, it represents the set of optimal trade-offs among competing objectives. 
To evaluate the quality of a Pareto front, the hypervolume metric (also called the S-metric) measures the portion of the objective space that is covered by the Pareto front in relation to a predefined reference point. 
This reference point is typically chosen to be worse than the ideal values across all objectives. 
The hypervolume is computed by determining the size of the region in the objective space that the Pareto front dominates while being constrained by the reference point.

\begin{figure}[h]
    \centering
    \includegraphics[width=0.8\linewidth]{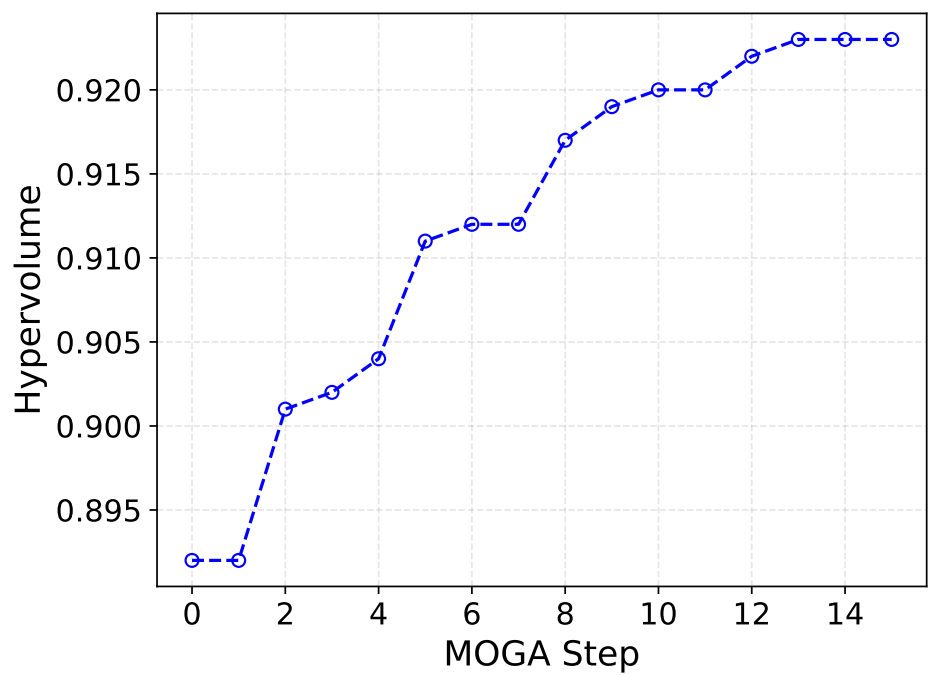}
    \caption{Hypervolume evolution during MOGA optimization}
    \label{fig:HV}
\end{figure}

MOGA was run for 15 steps and stopped after Pareto-hypervolume started to plateau as shown in Fig.~\ref{fig:HV}. 
We used the standard reference point of (0, 0) for the hypervolume calculation. 
The MOGA parameters of population size, mutation score, and cross-over probability were set to their default values of 64, 0.01, and 0.95, respectively. 
From all the possible constraint values on the Pareto front, we chose the one that achieved a minimum efficiency of 0.99 for highest purity. 
The efficiency and purity of the graph-builder with these constraints as listed in section~\ref{sec:Pre-Processing} are 0.9905 and 0.5473, respectively. 

